# Infant Brain MRI Segmentation with Dilated Convolution Pyramid Down-sampling and Self-attention

Zhihao Lei[1], Lin Qi[1], Ying Wei[1], Yunlong Zhou[1], Wenxu Qi[2]

[1] Northeastern University, Shenyang 110817, China
[2] Department of Radiology, Shengjing Hospital, China Medical University
Corresponding author: Ying Wei (weiying@ise.neu.edu.cn)

**Abstract.** In this paper, we propose a dual aggregation network to adaptively aggregate different information in infant brain MRI segmentation. More precisely, we added two modules based on 3D-UNet to better model information at different levels and locations. The dilated convolution pyramid down-sampling module is mainly to solve the problem of loss of spatial information on the down-sampling process, and it can effectively save details while reducing the resolution. The self-attention module can integrate the remote dependence on the feature maps in two dimensions of spatial and channel, effectively improving the representation ability and discriminating ability of the model. Our results are compared to the winners of iseg2017's first evaluation, the DICE ratio of WM and GM increased by 0.7%, and CSF is comparable. In the latest evaluation of the iseg-2019 cross-dataset challenge, we achieve the first place in the DICE of WM and GM, and the DICE of CSF is second.

## 1 Introduction

In brain research, brain tissue segmentation is a important part. The accurate segmentation of infant brain tissue MR images is the key to studying early brain development. In recent years, deep convolutional neural networks have shown great potential for medical image analysis. Many models based on FCN [1], U-Net [2], and DeepLab [3] have achieved remarkable results in the field of medical segmentation. Multitudes of scholars have made great efforts to accurately segment brain MR images of infants in isointense stage (approximately 6-8 months of age). Nonetheless, gray matter (GM) and white matter (WM) in brain tissue of infants in isointense stage show similar intensity in magnetic resonance images. At the same time, the brain images of infants also has the problem of blurred tissue boundaries, a lot of noise and too few samples, which all limit the further improvement of the performance of the algorithm.

The 3D-UNet network is a very powerful structure in medical imaging segmentation and previous work [4, 5, 6] has demonstrated that this kind of encoding-decoding network structure can perform better in brain tissue segmentation., but there are also two problems: (1) Continuous down-sampling in the encoding phase will cause spatial information loss, which results in the network unable to accurately locate the category of each voxel. (2) The network decoding phase cannot totally consider the global information, which is not conducive to the network to predict the pixel categories.

To solve the first problem, we propose an improved dilated convolution pyramid structure to form the down-sampling module. By using the dilated convolution [7] of different dilation rates, the features are extracted from multiple scales before down-sampling, avoiding the loss of detailed information on the down-sampling stage. Specifically, we use four different convolutions to construct four levels of features, which are then combined to form the final feature for down-sampling. This allows each down-sampling to be performed on four different features, which preserves different features of the next level of features. Our proposed dilated convolution pyramid down-sampling module can provide multi-scale features, so that the semantic information can be extracted in the encoding phase without a lot of details being lost.

For the second problem, we propose a 3D-based improved dual-channel attention module that combines channel attention module and spatial attention module, and ultimately combines the information that is focused on both by concatation operations. Unlike the previous work, we are using the attention mechanism on the 3D medical image and increasing channel attention which can effectively models long semantic dependencies, so we can use the edge information and details of the image more effectively.

The contribution of our work can be summarized as follows:

— We utilize the spatial convolution structure of the spatial pyramid based on the down-sampling of Res-Net, and utilize the convolution of different dilation rates to capture multi-scale information and can effectively expand the receptive field of the convolution kernel.
— A self-attention based on 3D medical images is proposed. The attention method we proposed can not only effectively encode the wider context information into local features, but also mine the interdependencies between channel maps and improve the feature representation of specific semantics.
— Our model has achieved very effective results through experimental demonstrations. In the iseg2019 cross-dataset competition, several indicators win the first place, proving that our model has significant generalization performance.

## 2 Proposed method

### 2.1 Algorithm framework

Our algorithm contains two parts: image pre-processing, model architecture. Our approach in the pre-processing and model architecture phases is as shown in Fig.1. In section 3.1, we will show the details of preprocessing. Due to the limited computational power of our 3-dimensional structure, in order to make our training process more efficient, we cropped the training sets and test sets to the size of 32*32*32, the overlapping step size is set to 8, then we feed them into the network.

## 2.2 Model architecture

In the convolutional layer of a deep network, the size of the convolution kernel determines the size of the receptive field, ie the size of the learning feature [8]. In the case of different convolution kernel sizes, features of different scales can be features are of great significance to improve the segmentation performance of deep networks, that is, multi-scale features. The proposed architecture is illustrated in Fig.1. If a large convolution kernel is used to obtain the receptive field, the training parameters will be too complicated and complicated, and the problem of over-fitting will be caused. For the conflict of large-scale features and computational cost, we utilize the dilated convolution. Dilated convolution is a special convolution operation that achieves large-scale features without adding too many parameters and computational costs. And we can get multi-scale features by changing the size of the dilation rate. In this paper, the use of dilated convolution in the encoding part is very helpful for extracting and aggregating multi-scale features, and favorable experimental results are obtained. In the encoder stage, we implement this function through the DCP module.

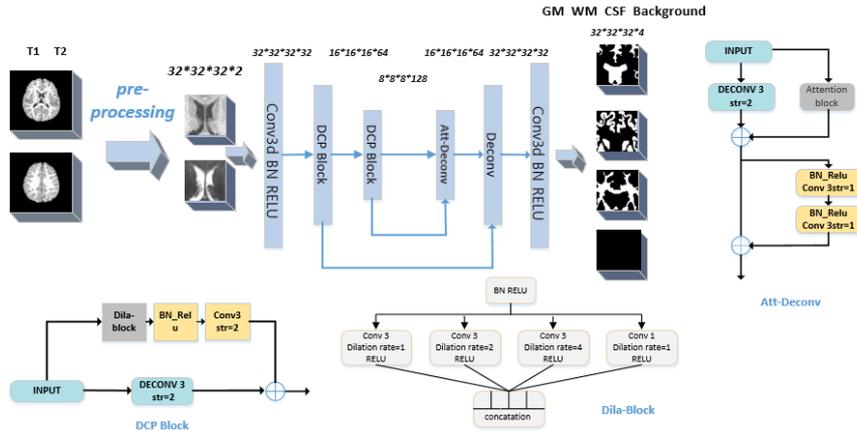

**Fig. 1.** Overview of our model architecture

In view of the low precision of the edge segmentation of brain tissue, we adopted the attention mechanism common in computer vision to enhance the edge significance area in the image, so as to make the output features more differentiated. We use channel attention and spatial attention in the decoder stage of the network, and the attention module can encode more extensive context information into local features in space, thus enhancing their representation ability. In the channel attention module, each high-level characteristic of channel map can be seen as a specific kind of response. By mining the interdependency relation between the channel maps, we can emphasize the interdependency feature map and improve the feature representation of specific semantics.

- DCP Block

Existing research [9] showed that the residual network can improve the training of deep learning models to some extent. We use a residual-network-based structure in down-sampling. We cancel the pooling layer and replace the commonly used pooling layer with a convolution with a stride of 2. The left side is a convolution of 1*1*1 with a stride of 2. The right side contains a dila-block module and an activated convolutional layer with a convolution kernel size of 3*3*3 and a stride of 2. About this dila-block. The structure is shown in the Fig.2, including a pre-activation layer with four dilated convolutions in the middle. The convolution kernel has a size of 3*3*3 and the first three dilation rates are 1, 2, 4, respectively, the last one is a 1*1*1 convolution. After the convolution, the four blocks are concated together and then connected to the convolution layer of the next stage.

By adopting this module, we can make full use of the shallow information of the image, and the pyramid structure also enables us to make full use of the multi-scale information.

- Attention Block

A common example is a self-attention block used in decoding phase that computes the output of a position by focusing on each position of the input. These studies [10, 11, 12] were all used to capture long-range dependencies, that is, to aggregate the information of feature graphs.

We also add a self-focus module to the decoder part of the 3D-Unet network, similar to DA-Net [13]. The difference is that we add a self-attention mechanism module to the 3D image. Most of the previous studies were based on two-dimensional research. Compared to two-dimensional, we need to process the 3D images into 2-dimensional vectors when processing the data. The structure is shown in Fig.2.

We first represent a $\boldsymbol{D*W*H}$ cube of the 3D image as a vector of $\boldsymbol{D*W*H*C}$, after the flatten operation, these vectors are folded into a matrix of size of (D*H*W)*C. For each vector generated by 3D cube B, we multiply it by the transpose of C, then get a feature map F through a soft-max. F is defined as:

$$\boldsymbol{F} = Softmax(\frac{\boldsymbol{BC^T}}{\sqrt{C_E}}) \qquad (1)$$

where $\boldsymbol{F} \in R^{(D*H*W)*(D*H*W)}$. We multiply the feature obtained map to finally obtain the vector V of the same size as the B generation vector.

$$\boldsymbol{V} = Q_D * \boldsymbol{F} \qquad (2)$$

After we have calculated each vector generated by (2), then we superimpose them together to produce the final output. In order to better integrate the sub-module into the network, we pass a 1*1*1 convolutional layer and then send it to the network.

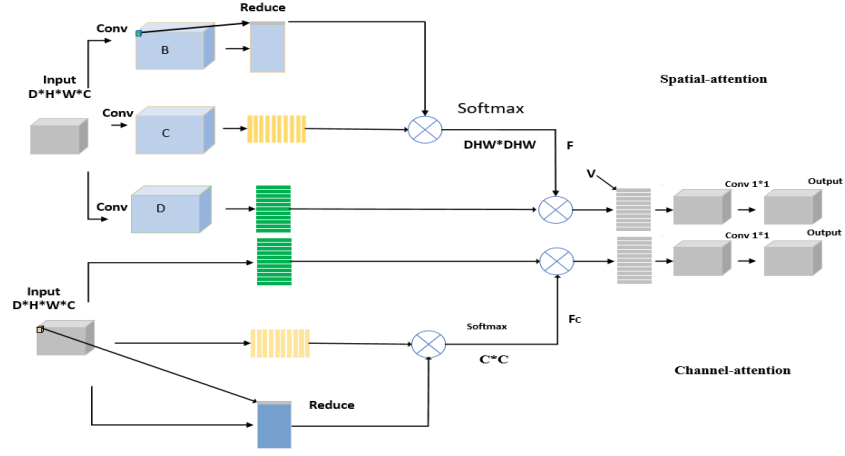

**Fig. 2.** Attention block

Finally, we get the same output as the original image:

$$\boldsymbol{O} = Conv(restore()) \tag{3}$$

where restore() is the reverse operation of the previous reduce operation. This module enhances their presentation capabilities by encoding a wider range of contextual information into local features.

 Channel Attention Module: As mentioned in the DA-Net, the channel map for each high-level feature can be thought of as a class-specific response, with different semantic responses associated with each other. We can mine the interdependencies between channel maps to emphasize interdependent feature mapping. Therefore, we have built a channel attention module to explicitly model the dependencies between channels.

$$\boldsymbol{K_C} = Q_A * \boldsymbol{F_C} \tag{4}$$

which is the same size as the image after flatten, and finally we perform the same operation as the positional attention to restore the original image size. The final feature of each channel is the weighted sum of all channel features and the original features, modeling the long semantic dependencies between the feature maps, which helps to improve feature discrimination.

 Finally, we connect the dual-attention module in parallel and use it in the decoder part. This module effectively utilizes the spatial attention and channel attention information to model the long-term semantic dependence of the feature map, which effectively improves the representation ability and discriminative ability of the model.

## 3   Experiment

### 3.1   Parameters setting

The optimization of network parameters is performed via Adam optimizer. Learning rate was initialized as 3.3 *10 -3 and we set weight decay as 2e-6 per 1000 iterations. the patch-size size was set to 32*32*32, and the batch-size size was 5. We have referenced other parameter settings based on paper. Some of these parameters have proven to be the best choice through experiments [14]. Because there is very little training data available，In the test process, we used the tenth sample as our verification sample. Experiments were performed in a computational server with one NVIDIA 1080Ti with 11GB of RAM memory. Each training session cost about 3 hours. It took about 2 minutes to segment a 3D MRI image on NVIDIA 1080Ti. Our evaluation indicators use DICE and ASD.

### 3.2   Data sets and results

In this section, we evaluate our results through two public data sets:

Iseg2017[1] [15] is from the Baby Connection Project, where UNC and UMN researchers conduct a four-year safe, non-invasive multimodality for 500 normally-developed children aged 0-5 years. Brain magnetic resonance imaging (MRI) scans (including T1 and T2 weighted structures MRI, DTI, and rsf-MRI) were used, and T1 and T2 MR images of 10 healthy 6-month-old infants were used. A total of 10 training sets and 13 test sets are included.

Due to the excellent features of 3d-unet, we use 3d-unet as our baseline. During model training, because there are fewer samples in the training set, we utilize the leave-one-subject-out cross validation to evaluate segmentation performance and void bias. For each training we use one sample as our validation set and the remaining nine as our training sets. In this way we evaluate the effectiveness of our model.

The experimental results are shown in table 1. Compared with the baseline and several other published papers, our GM, WM results have some big improvements, which is 0.7% higher than the previous best results.

In the iseg2019 cross-data set[2], the test set contains samples of three other data sets, which are provided by BCP, Stanford University, Emory University. For preprocessing, these images have been resampled into $1 \times 1 \times 1$ mm3 to eliminate the effect of resolution, and employed the same tools for skull stripping and intensity inhomogeneity correction.

The results of our final segmentation were evaluated by the contest organizer. So far, there are no published papers can be compared, the criterion we use to evaluate is to compare with other participants. The experimental results are shown in table 2. Of the 9 indicators, ranking first in five indicators of GM and WM, and the gap between us and the first in CSF is very small. Moreover, we have an obvious lead in WM seg-mentation results, the DICE value is 1.4% higher than the second. The challenge's rank can be found in: `http://iseg2019.web.unc.edu/evaluation-results/`.

---

[1] See http://iseg2017.web.unc.edu/
[2] See http://iseg2019.web.unc.edu/

Table 1. Comparison of our results with other methods.

| Method | CSF | | WM | | GM | |
|---|---|---|---|---|---|---|
| | DSC | ASD | DSC | ASD | DSC | ASD |
| Msl_SKKU | 95.8 | 0.116 | 90.1 | 0.391 | 91.9 | 0.330 |
| LIVIA[16] | 95.7 | 0.138 | 89.7 | 0.376 | 91.9 | 0.338 |
| Bern_IPMI | 95.4 | 0.127 | 89.6 | 0.398 | 91.6 | 0.341 |
| Hyper[17] | 95.6 | 0.120 | 90.1 | 0.382 | 92.0 | 0.329 |
| Baseline | 92.3 | 0.142 | 84.6 | 0.412 | 89.8 | 0.365 |
| Ours | 95.9 | 0.114 | 90.8 | 0.353 | 92.6 | 0.307 |

Table 2. The results of cross-dataset segmentation.

| Method | CSF | | WM | | GM | |
|---|---|---|---|---|---|---|
| | DSC | ASD | DSC | ASD | DSC | ASD |
| TAO | 83.5 | 0.526 | 86.0 | 0.515 | 83.0 | 0.543 |
| Xflz | 82.6 | 0.556 | 86.1 | 0.522 | 82.5 | 0.558 |
| Fight | 82.6 | 0.570 | 86.3 | 0.523 | 82.9 | 0.561 |
| WYF | 82.9 | 0.546 | 85.0 | 0.547 | 82.6 | 0.568 |
| Ours | 83.4 | 0.553 | 87.7 | 0.474 | 83.5 | 0.513 |

### 3.3 Ablation experiments

The result of our baseline is the average of 10 samples. Results of our model of CSF and GM on the DICE are 3% higher than the base model, and the WM is increased by 5.4%. At the same time, in order to explore the promotion of each module. We perform the following 2 ablation experiments:

Model-1: In this ablation experiment, we validate the effectiveness of the proposed self-attention model, which remove the self-attention module based on our model.

Model-2: The down-sampling part is removed from the proposed model to verify its effectiveness.

From the Table 3, we can see that each module of our proposed model is very helpful for our final result, and it has a big improvement compared with the baseline.

Table 3. Ablation experiment results.

| Model | CSF | WM | GM | AVG |
|---|---|---|---|---|
| Model-1 | 95.3 | 87.0 | 91.6 | 91.6 |
| Model-2 | 94.6 | 86.3 | 90.8 | 90.6 |
| Baseline | 92.3 | 84.6 | 89.8 | 88.3 |
| Our model | 95.3 | 91.3 | 92.4 | 92.9 |

Fig.3 shows the results of our final segmentation visualization in iseg2017. From left to right are ground truth, baseline and our model respectively. The three upper pic-tures shows the results of WM on the gold standard, baseline, and our model. We clearly see that the edge of the baseline is not smooth and has many parts outside the boundary, while our model has achieved relatively excellent results in most areas (in red circle). Through the images below, we can see the baseline in the GM is not clear in gyrus structure. In contrast, the results of our model can clearly see the gyrus (in blue circle), which is inseparable for the improvement of DICE and ASD values.

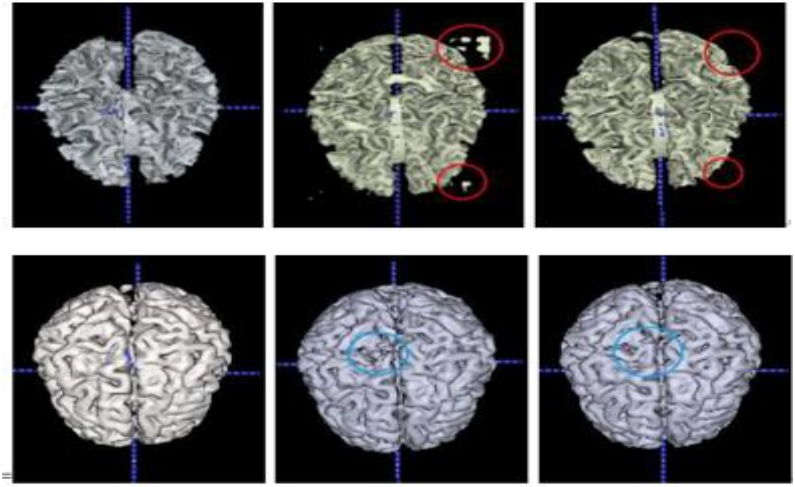

**Fig. 3.** GM and WM results of our method

## 4    Conclusion

In this paper, we present a dual aggregation network based on pyramid structure and self-attention mechanism in multimodality MRI segmentation of infants. Existing 3D-Unet segmentation methods can easily cause global information loss during the down-sampling phase. In the decoder part, the deconvolution cannot recover all necessary information during the up-sampling process, which reduces the accuracy of image segmentation. Our proposed spatial pyramid-based down-sampling and self-attention mechanisms minimize image loss during image convolution. At the same time, our model also has excellent performance in the segmentation of cross-dataset competition, indicating its excellent generalization performance. It can be said that our model has excellent performance in brain tissue segmentation tasks.

**Acknowledgement.**

Zhihao Lei and Lin Qi contributed equally to this work. This work is supported by National Natural Science Foundation of China (No. 61871106).